\documentclass[conference]{IEEEtran}
\IEEEoverridecommandlockouts

\usepackage{cite}

\usepackage[flushleft]{threeparttable}

\ifCLASSINFOpdf
\else
\fi
\usepackage{url}
\usepackage{hyperref}

\usepackage{graphicx}
\usepackage{acronym}
\usepackage[utf8]{inputenc}
\usepackage{url}

\usepackage{todonotes}

\newacro{AB}[AB]{AdaBoost}
\newacro{DT}[DT]{Decision Tree}
\newacro{F3}[F3]{Fake Feature Framework}
\newacro{GNB}[GNB]{Gaussian Naive Bayes}
\newacro{LDA}[LDA]{Latent Dirichlet Allocation}
\newacro{LR}[LR]{Logistic Regression}
\newacro{P}[P]{Personal Features}
\newacro{POS}[POS]{Part of Speech}
\newacro{RA}[RA]{Reviewing Activity Features}
\newacro{RF}[RF]{Random Forest}
\newacro{S}[S]{Social Features}
\newacro{T}[T]{Trusting Features}
\newacro{TF-IDF}[TF-IDF]{Term Frequency - Inverse document frequency}
\newacro{WOM}[WOM]{word-of-mouth}

\begin{document}
%
\title{A Framework for Fake Review Detection\\ in Online Consumer Electronics Retailers\\ }

\author{\IEEEauthorblockN{Rodrigo Barbado, Oscar Araque, Carlos A. Iglesias}
\IEEEauthorblockA{Intelligent Systems Group\\
Department of Telematic Engineering Systems\\
Universidad Politécnica de Madrid\\
ETSI Telecomunicación, Avda. Complutense, 30, 20840 Madrid, Spain\\
Email: rodrigo.barbado.esteban@alumnos.upm.es, o.araque@upm.es, carlosangel.iglesias@upm.es}
}


%


\onecolumn
\maketitle

\begin{abstract}
The impact of online reviews on businesses has grown significantly during last years, being crucial  to determine business success in a wide array of sectors, ranging from restaurants, hotels to e-commerce. Unfortunately, some users use unethical means to improve their online reputation by writing fake reviews of their businesses or competitors. Previous research has addressed fake review detection in a number of domains, such as product or business reviews in restaurants and hotels. However, in spite of its economical interest, the domain of consumer electronics businesses has not yet been thoroughly studied. This article proposes a feature framework for detecting fake reviews that has been evaluated in the consumer electronics domain. The contributions are fourfold:  (i) construction of a dataset for classifying fake reviews in the consumer electronics domain in four different cities  based on scraping techniques; (ii) definition of a feature framework for fake review detection; (iii) development of a fake review classification method based on the proposed framework and (iv) evaluation and analysis of the results for each of the cities under study. We have reached an 82\% F-Score on the classification task and the Ada Boost classifier has been proven to be the best one by statistical means according to the Friedman test. 



\end{abstract}



%
\IEEEpeerreviewmaketitle

\section{Introduction}
Online consumer product reviews are playing an increasingly important role for customers, constituting a new type of\ac{WOM} information~\cite{chen2008online}. Recent research shows that 52\% of online consumers use the Internet
to search for product information, while 24\% of them use
the Internet to browse products before making purchases~\cite{ha2015impact}. As a result, online reviews has a strong impact on consumers' decision purchase in e-commerce, affecting the most relevant areas, such as travel and accommodations~\cite{filieri2014wom,sotiriadis2013electronic}, online retailers~\cite{awad2008establishing}, and entertainment~\cite{zhu2006influence,chevalier2006effect,dhar2009does}. Moreover, online reviews of the same product can be found in multiples sources of information, which can
be classified~\cite{park2012relationship} according to the parties that host \ac{WOM} information into internal \acp{WOM}, hosted by retailers (e.g. Amazon, Walmart, BestBuy, etc.) and external ones, hosted by independent product review providers (e.g. CNET, Yelp, TripAdvisor, Epinions, etc.). 

Nevertheless, only credible reviews have a significant impact on consumers' purchase decision~\cite{chakraborty2018effects}.
Moreover, product category affects significantly the credibility of \acp{WOM}~\cite{mudambi2010research}. Consumer electronics product category is the most online reviewed~\cite{chan2011conceptualising}, based on a number of factors. On the one hand, consumer electronics usually require a significant investment, and the more valuable and expensive an item is, the more it is researched. 
According to a study~\cite{riegner2007word}, consumer electronics are the product most influenced by online reviews, influencing the 24\% of products acquired in this category, and being \acp{WOM} the second most influential source after search engines in this product category.  On the other hand, consumers tend to research on consumer electronics products because these products change very frequently, with new products and updates of existing ones~\cite{chakraborty2018effects}. Thus, consumers frequently trust on reviews to avoid making a wrong purchase decision~\cite{park2008effects}. As a result, Horrigan et al.~\cite{horrigan2008internet} report that more than 50\% of consumer electronics buyers tend to consult several \acp{WOM} before making a purchase decision.

Some studies~\cite{gu2012} show that retailer hosted online \ac{WOM} influences enormously sales in low involvement products, such as books or CDs. However, consumers usually  conduct a pre-sales research in high-involvement products, such as consumer electronics. Thus, in consumer electronics, retailer's internal \ac{WOM} has a limited influence, while external \ac{WOM} sources have a significant impact on the retailer's reputation and sales~\cite{cui2012effect}.  Hence, consumer electronics are more sensible to the effects of external \acp{WOM}, since they cannot easily act on them.  


Since both consumers and retailers  become overwhelmed by the huge number of available opinions in \ac{WOM} internal and external sources, automatic natural language processing and sentiment analysis  techniques have been frequently applied. Some of the most frequent application domains are review polarity classification~\cite{poria2016aspect},  review summarization~\cite{potthast2010opinion}, competitive intelligence acquisition~\cite{dey2011acquiring} and reputation monitoring~\cite{ziegler2006towards}.



Given the importance of reviews for businesses and the difficulty of obtaining a good reputation on the Internet, several techniques have been used to improve online presence, including unethical ones.
Fake reviews are one of the most popular unethical methods which are present on sites such as Yelp or TripAdvisor. However, according to Jindal and Liu~\cite{jindalandliu1}, not all fake reviews are equally harmful. Fake negative reviews on good quality products are really harmful for enterprises, and along with fake positive reviews on poor quality products, result also harmful for consumers. Fake positive reviews on poor quality products are also harmful for competitors who offer average or good quality products but do not have so many reviews on them. 


The goal of this article is analyzing the fake review problem in the consumer electronics field, more precisely studying Yelp businesses from four of the biggest cities of the USA. No prior research has been carried out in this concrete field, being restaurants and hotels the most previously studied cases. We want to prove that fake review detection problem in online consumer electronics retailers can be solved by machine learning means and to show if the difficulty of achieving it depends on geographic location.


In order to achieve this goal, we have followed a principled approach.
Based on literature review and experimentation, a feature framework for fake review detection is proposed, which includes some contributions such as the exploitation of the social perspective.
This framework, so called \acf{F3}, helps to organize and characterize features for fake review selection.
\acs{F3} considers information coming from both the user (personal profile, reviewing activity, trusting information and social interactions) and review elements (review text), establishing a framework with which categorize existing research.
In order to evaluate the effectiveness of the features defined in \ac{F3}, a dataset from the social Yelp in four different cities has been collected and a classification model has been developed and evaluated.





The reminder of the paper is structured as follows. Sect.~\ref{sec:related-work} reviews the state of the art on fake review detection on other domains.
Afterwards, Sect.~\ref{sec:methodology} presents the followed methodology and also introduces the proposed feature framework.
Experimentation is detailed in Sect.~\ref{sec:evaluation}. 
Finally, Sect.~\ref{sec:conclusions} highlights and discusses the main obtained results.

\begin{table}
 \begin{center}
	\begin{tabular}{ c  c  c  c  c  c  c  c  c  c }
	\hline
	Reference & Year & Domain & Algorithms & Personal & Social & Review Activity & Trust & Review centric \\ \hline
	Liu et al.~\cite{liu2017identifying}& 2017& E-Commerce & -&  &  x & x & x & x &   \\ \hline
	Zhang et al.~\cite{zhang2016}& 2016& Hotel   & SVM, NB, DT, RF, LR  & x & x & x & x & x &   \\ \hline
	Li et al.~\cite{li2016detecting}& 2016& Restaurant & SVM, NB, DT  &  & x  & x  & x & x  &   \\ \hline
	Heydari et al.~\cite{heydari2016detection}& 2016  & E-Commerce  & - &  & x  & x &  &   \\ \hline
	Luca et al.~\cite{luca2016fake}& 2016& Restaurant   & -  & x & x & x  &  & x  &   \\ \hline
	Dewang et al.~\cite{dewang2015identification}& 2015& Hotel & NB  &   &   &  &  & x &   \\ \hline
	Li et al.~\cite{li2015analyzing}& 2015& Restaurant & SVM  &   &   & x &  & x &   \\ \hline
	Banerjee et al.~\cite{banerjee2015using}& 2015& Hotel & SVM, NB, RF, DT, LR  &   &   &  &  & x &   \\ \hline
	Hernández et al.~\cite{HERNANDEZFUSILIER2015433}& 2015& Hotel   & SVM, NB  &  &  &  &  & x &   \\ \hline
	Fornaciari et al.~\cite{fornaciari2014identifying} & 2014 & E-Commerce & SVM & & & & x & \\ \hline
	Akoglu et al.~\cite{akoglu2013opinion}& 2013& Software   & -  &  &  &  & x &  &   \\ \hline
	Mukherjee et al.~\cite{mukherjee1}& 2013& Hotel, Restaurant   & SVM  &  &  & x & x & x &   \\ \hline
	Fei et al.~\cite{fei2013exploiting}& 2013& Software & SVM  &  &  & x & x &  &   \\ \hline
	Li et al.~\cite{lietal}& 2011& E-Commerce   & NB  & x & x  & x & x  & x &   \\ \hline
	Ott et al.~\cite{ottetal}& 2011& Hotels   & SVM, NB  &  &  &  &  & x &   \\ \hline
	Wang et al.~\cite{6137345}& 2011& E-Commerce   & -  &  &  &  & x &  &   \\ \hline
	Jindal et al.~\cite{jindalandliu2} & 2008& E-Commerce & SVM &   &   & x & x & x &   \\ \hline

	\end{tabular}
	   \caption{Reviewed works classified according to \acs{F3} framework}
	\label{table:stateframework}
\end{center}
\end{table}

\section{Related work} \label{sec:related-work}
The task of fake review detection has been studied since 2007, with the analysis of review spamming~\cite{jindalandliu1}.
In this work, the authors analyzed the case of Amazon, concluding that manually labeling fake reviews may result challenging, as fake reviewers could carefully craft their reviews in order to make them more reliable for other users.
Consequently, they proposed the use of duplicates or nearly-duplicates as spam in order to develop a model that detects fake reviews~\cite{jindalandliu1}. Research on distributional footprints has also been carried out, showing a connection between distribution anomalies and deceptive reviews from Amazon products and TripAdvisor hotels~\cite{feng2012distributional}.




Fake review detection is a specific application of the general problem of deception detection, where both verbal and nonverbal clues can be used~\cite{fitzpatrick2015automatic}. Fake review detection research has mainly exploited textual and behavioral features, while other approaches have taken into account social or temporal aspects. 

Textual features have been proposed in several papers. Ott et al.~\cite{ottetal} employed psycholinguistic features based on LIWC~\cite{tausczik2010psychological} combined with standard word and \ac{POS} n-gram features. Mukherjee et al. \cite{mukherjee1} extend that work including also style and \ac{POS} based features, such as deep syntax and \ac{POS} sequence patterns.  However, the detection of fake reviews based only on textual features is challenging. Other articles propose additional textual features such as 
semantic similarity and emotion~\cite{li2016detecting}, a wide variety of lexical and syntactic features~\cite{dewang2015identification} and deeper details such as understandability, level of details, writing style and cognition indicators~\cite{banerjee2015using}.


Behavioral features refer to nonverbal characteristics of review activity, such as the number of reviews or the time and device where the review was posted. They were used in order to improve the classification model resulting in encouraging results. Liu et al.~\cite{jindalandliu2} introduced behavioral features on Amazon reviews, distinguising among review features (e.g. number of feedbacks, position of the review,  textual features, rating features, etc.),  product features (e.g. price, sales rank) and reviewer features (e.g. average rating, ratio of the number of reviews that the reviewer wrote which were the first reviews, etc.). In another work, Zhang et al.~\cite{zhang2016} explore the effect of both textual and behavioural features in the restaurant and hotel domain, showing that non-textual features result more relevant for the task of fake review detection. Also, regarding the restaurant domain, some interesting findings were described by Luca et al.~\cite{luca2016fake}. Restaurants are more likely to make review fraud when they have a lower reputation, including having few reviewers or bad scorings.

Apart from using textual and behavioral features, other methodologies were followed for the fake review detection task. Wang et al.~\cite{6137345} proposed a review graph with the aim of capturing relationships between reviewers, reviews and stores reviewed by the reviewers. Making use of this graph, an iterative model was used to identify suspicious reviewers. Following also a graph model, network effects were analyzed by Akoglu et al.~\cite{akoglu2013opinion}, following two steps: user and review scoring for fraud detection and grouping for visualization. 

Another methodological approach focuses on temporal aspects, and concerns the burstiness of reviews and their impact on businesses. Bursts of reviews can be either due to sudden popularity of products or spam attacks~\cite{fei2013exploiting}, which were also analyzed in \cite{liu2017identifying} along with other behavioral and textual features. A deeper time series approach was made by Heydari et al.~\cite{heydari2016detection} and Li et al.~\cite{li2016detecting} propose other types of features such as review density in temporal windows, along with semantic and emotion features. Spatial and temporal features were used in a Chinese site by Li et al.~\cite{li2015analyzing}.


Regarding classification algorithms, Support Vector Machine~\cite{vapnik1997support} was the most used one followed by Naive Bayes~\cite{friedman1997bayesian}, Decision Tree~\cite{breiman1984classification}, Random Forest~\cite{breiman2001random} and Logistic Regression~\cite{cox1958regression} as shown in Table~\ref{table:stateframework}.


Apart from supervised learning, other approaches have been followed, since collecting data for experiments is a hard task.
In~\cite{lietal}, authors propose a prediction model based on semi-supervised learning and a set of textual and behavioural features.
Additionally, Hernandez et al.~\cite{HERNANDEZFUSILIER2015433} propose a semi-supervised technique called PU-learning.

The described prior research highlighted in this section has been organized  according to our proposed framework as shown in Table~\ref{table:stateframework}. Regarding the field of application, this table shows that previous research has been centered around restaurant and hotel domains when considering reviews about businesses. To the extent of our knowledge, the study of the problem of fake review detection on the consumer electronics domain is a novel work. Moreover,  its application is highly relevant given that consumer electronics are more sensible to an unethical \ac{WOM} misuse.



\begin{figure*}
 \centering
 \includegraphics[width=0.9\textwidth]{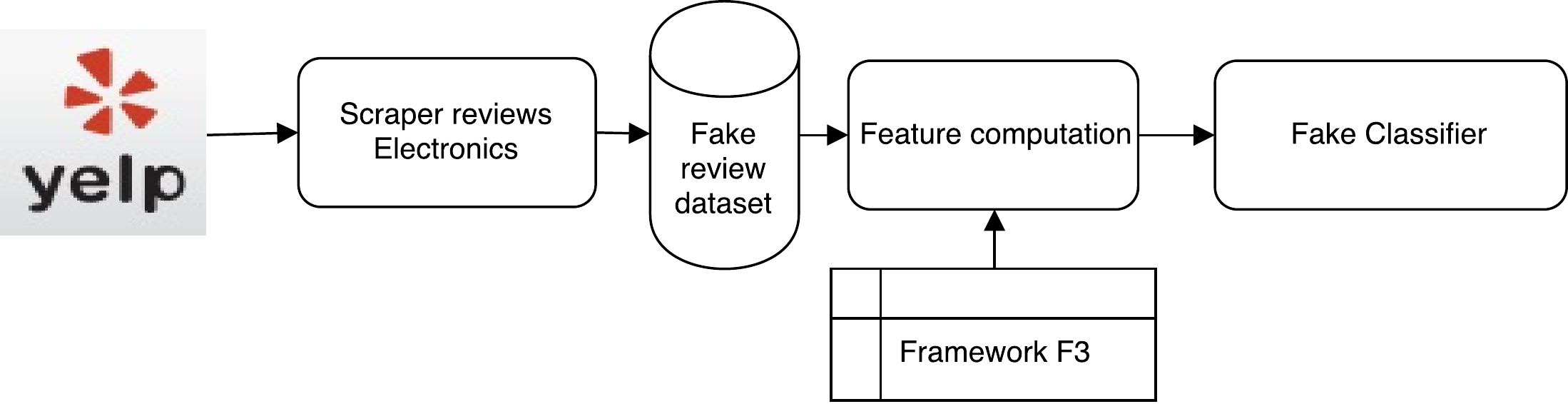}
 \caption{Methodology}
 \label{fig:methodology}
\end{figure*}

\section{Methodology}\label{sec:methodology}

The methodology followed in this article is shown in Fig.~\ref{fig:methodology}. The first step is building the dataset from Yelp by web scraping means (Sect.~\ref{sec:scraping}). Then, a feature model is defined and computed (Sect.~\ref{sec:f3}) for training a classifier that detects fake reviews (Sect.~\ref{sec:evaluation}).

\subsection{Scraping process}\label{sec:scraping}

As the consumer electronics field has not been studied before, there is not an available dataset to experiment with, so the
starting point consists on data collection. 
Furthermore, Yelp's filter has been used as a reference for labelling reviews as fake or not since, according to its CEO, Yelp's filtering algorithm has evolved over the years to filter fake reviews~\cite{mukherjee1}.
Also, this filter has been claimed to be highly accurate~\cite{weise2011lie}.

As Yelp shows both trustful and filtered reviews, it is possible not only to extract information about reviews and users but also whether if they had been filtered or not. So, the first step was developing a web scraper to gather the necessary information for the experiments. This web scraper was programmed in Python using Scrapy~\cite{kouzis2016learning}, a library which offers the possibility of building web crawlers. The resulting corpus was compound by reviews from four important cities of the USA: New York, San Francisco, Los Angeles and Miami.
Yelp offers the possibility of searching category businesses in each city, so the scraping process is focused on the pages in which all electronic businesses from the different selected cities appeared. 
For each of these pages, all businesses appearing on it were selected in order to retrieve their information.



Secondly, for each business site all the reviews appearing on it were collected along with their user profile URL. In this step, reviews were also labeled as fake or not according to the Yelp filter.

The last step of the corpus creation consisted on gathering all the information appearing on each user profile site through the URLs which were collected in the previous step. Each of those profile sites show all the necessary information to develop the features of the proposed framework.


As a result of the scraping process, the dataset shown in Table~\ref{table:dataset} has been extracted.
As expected, the distribution of fake and trustful reviews is highly unbalanced. 
In~\cite{mukherjee1} it was shown that around 14\% of reviews were fake. For our experiments, we crawled the same amount of reviews for each category, thus obtaining a balanced dataset. 
The dataset contains labeled reviews but also social aspects related to the user profile and her networking activity. 

\begin{table}[hbtp]
 \begin{center}
	\begin{tabular}{ l  l  l }
	\hline
	City & Trustful Reviews & Fake Reviews \\ \hline
	New York & 2472 & 2472 \\ 
	Los Angeles & 3776 & 3776 \\ 
	Miami & 1409 & 1409 \\ 
	San Francisco & 1799 & 1799 \\ \hline
	\end{tabular}
	\caption{Sizes of reviews per city of the corpus}
	\label{table:dataset}
\end{center}
\end{table}
 

An exploratory data analysis of the  dataset reveals interesting information. Table~\ref{table:words} shows  the most frequent words, which are similar between all cities. Moreover, and what is more important for the fake review detection task, there are not significant differences of popular words among trustful and fake reviews. This fact indicates that text features could not be very relevant for fake review discrimination.

\begin{table}[hbtp]
 \begin{center}
	\begin{tabular}{ l  l  l }
	\hline
	City & Trust & Fake  \\ \hline
	New York & Phone & Phone  \\
	         & Store & Service  \\ 
	         & Service & Store   \\ 
	         & Screen & Back  \\ 
	         & Time & Customer  \\ \hline
	         
	Los Angeles & Service & Service   \\ 
	            & Phone  & Phone   \\ 
	            & Store & Store  \\
	            & Great & Great   \\
	            & Place & Customer   \\ \hline
	Miami & Phone & Phone  \\
	      & Service & Service  \\
	      & Store & Store \\ 
	      & Customer & Great \\ 
	      & Screen & Customer  \\ \hline
	San Francisco & Phone & Phone \\ 
	              & Service & Service  \\
	              & Time & Store  \\
	              & Store & Customer  \\
	              & One & Time\\ \hline
	\end{tabular}
    \caption{Five most frequent words by city and class}
	\label{table:words}
\end{center}
\end{table}



On the other hand, the user information obtained from our web scraper shows clear unequal statistical values between fake and trustful users. This information is presented in Table~\ref{table:userFeatures},  showing the mean value, standard deviation and maximum values for each of the fields extracted from the user's profile site from Yelp (minimum values were not included as they were always zero). The maximum values are shown for informative purposes, as they were afterwards normalized. It can be observed that  fake reviewers tend to give lower ratings on their reviews than trustful reviewers, being their mean values 1.1 stars and 2.79 stars respectively from a maximum of 5.

\begin{figure*}[hbtp]
 \centering
 \includegraphics[width=\linewidth]{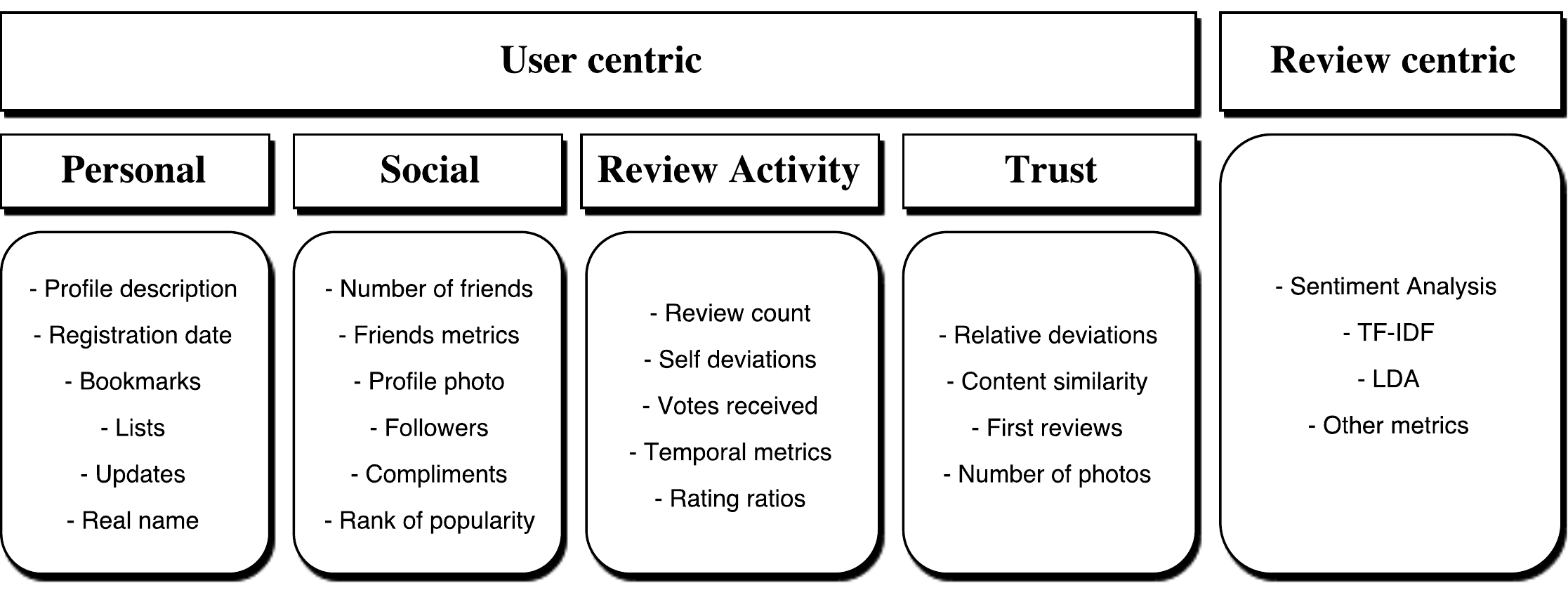}
 \caption{Overview of the  \acf{F3}}
 \label{fig:mesh1}
\end{figure*}

\subsection{\acf{F3}}\label{sec:f3}

In this section we introduce a feature framework \acf{F3} for organizing the extraction and characterization of features in fake detection. Its definition is inspired on the analysis of previous research, and includes a  novel definition of social features. Previous works have classified features into textual, behavioural and product features~\cite{jindalandliu2}. 
Our main contribution consists in providing a more detailed classification of user centric features, taking advantage of the social aspects of a social network such as Yelp.

As shown in Fig.~\ref{fig:mesh1},  our framework distinguishes review centric and user centric features. 


{\bf User centric features} consider information related to how users behave in a social network such as Yelp and which information users provide; and are divided into four types: {\bf \ac{P}}, {\bf \ac{S}}, {\bf \ac{RA}}, and {\bf \ac{T}}.

The first type, {\bf \ac{P}}, is  information related to user profile, such as the self description written by the user; users' businesses subscriptions, known as bookmarks; lists containing several bookmarks; registration date; updates made on self reviews; and the user's real name.
This information is available in social review networks, such as Yelp, and can be automatically obtained.

The second type, {\bf \ac{S}}, is the feature set related to the way the user interacts with other users. In our case, we have identified these features by inspecting the Yelp social features. Our hypothesis is that social activity can help in the classification task, since social features can help to extend the context of the linguistics features. Several works have also proposed this approach in a number of applications, such as sentiment analysis~\cite{tan2011user} or stance detection~\cite{lai2017extracting}. Social features included here are number of friends, metrics gathered from friends such as their number of reviews or friends, number of followers, number of compliments, rank of popularity and the presence of a profile photo. 

The third type, {\bf \ac{RA}} is the feature set related to the way the user posts his reviews. Some examples of reviewing activity features are review counts~\cite{mukherjee1}, self review deviations \cite{jindal2007analyzing}, number of received votes (in Yelp there are three types: `cool', `funny' and `useful'), number of posted tips (which are shorter reviews that appear on Yelp), maximum number of reviews in a date or specific review counts in temporal windows. Additionally, metrics related to the ratios of positive and negative reviews are included here.

A last type, {\bf \ac{T}}, is the feature set that aims at pointing out inconsistencies or abnormal behaviour in the user review activity. For example, a high content similarity of the reviews of a given user could reveal that said user may be using templates for reviewing \cite{mukherjee1}. In this regard, rating deviations with respect to other users reviewing the same businesses~\cite{jindal2007analyzing} can also offer greater insight. Other features included in this type are the number of first reviews and number of uploaded photos.

In the case of {\bf review centric features}, apart from textual metrics such as average lengths or using techniques such as \ac{TF-IDF}~\cite{OUYANG2011227}, we have also included the use of \ac{POS} tags~\cite{PETZ2014899}, \ac{LDA} models~\cite{CHEN20171299}, Word2Vec models~\cite{goldberg2014word2vec} and sentiment analysis for enriching the available features using the information we have. Emotion analysis has also been used in prior research as well as other lexical and syntactic features.

This framework has been used for classifying state-of-the-art features as shown in Table~\ref{table:stateframework}.  

After experimenting with different potential interesting features, the features shown in Table~\ref{table:userFeatures} have been selected.


 \begin{table*}[hbtp]
 \begin{center}
	\begin{tabular}{ l | l | l l l | l l l }
	\hline
	Subset & Feature & & Trust & & & Fake &   \\ \hline
	        && mean & std & max & mean & std & max  \\
    	&Profile description & 0.19& 0.39& 1.0 & 0.06& 0.24&1.0    \\ 
	Personal    &Bookmark lists & 36.47 & 183.19&  5842.0& 2.09 & 27.74 & 1717.0  \\
	    &Lists & 1.45 & 15.67&  712.0 & 0.04 & 0.58& 30.0   \\ 
	    & Review updates & 4.22 & 20.80&  562.0 & 0.34 & 2.52& 85.0   \\ \hline
	 &Friends' no. of friends & 231.75&417.09& 5000.0 & 66.77 & 269.39  & 13699.0   \\ 
	&Friends' no. of reviews & 80.6 & 189.99 & 2603.0& 26.70 & 121.96& 2885.0   \\ 
	&Profile has photo & 0.76 & 0.43& 1.0 & 0.41& 0.49& 1.0   \\ 
	Social &No. of followers & 6.18& 45.34& 1782.0 & 0.38 & 5.02& 263.0   \\ 
	& No. of friends & 70.86 & 260.70& 5000.0 & 13.90 & 106.14& 5000.0    \\ 
	&No. of votes `cool' & 155.91&1169.02& 35842.0& 5.41& 112.61& 5440.0  \\ 
	
	&No. of votes `useful' & 231.35& 1449.30& 51012.0 & 8.58& 128.43&6170..0    \\ 
	&No. of votes `funny' & 136.18 & 1010.25&  32844.0& 4.35 & 92.27 & 4184.0  \\ \hline
	&No. of reviews & 77.71 & 328.41& 11225.0 & 7.78 & 42.14& 1404.0   \\ 
	& Rating distribution (\% 5 stars) & 0.37 & 0.31& 1 & 0.14 & 0.27& 1.0   \\ 
	& Rating distribution (\% 4 stars) & 0.13&0.16& 0.83 & 0.05 & 0.13 & 1   \\ 
	Review Activity & Rating distribution (\% 3 stars) & 0.06 & 0.09& 1.0& 0.02 & 0.07& 0.8   \\ 
	& Rating distribution (\% 2 stars) & 0.05 & 0.08& 0.8 & 0.02& 0.06& 0.6   \\ 
	& Rating distribution (\% 1 star) & 0.12& 0.17&1.0 & 0.07 & 0.18& 1.0  \\ 
	&Average rating & 2.79& 1.78&5.0 & 1.1 & 1.74& 5.0  \\ \hline
	Trust &No.  of photos & 127.39 & 1135.01& 57761.0 & 5.60 & 141.04& 7599.0    \\ 
	&No. of tips & 24.29&269.99& 16364.0& 1.27& 18.56& 1040.0  \\ \hline
    
	\end{tabular}
    \caption{Selected features distributions over the whole dataset}
	\label{table:userFeatures}
\end{center}
\end{table*}



\section{Experimentation and evaluation} \label{sec:evaluation}

This section describes the experiments carried out to develop and evaluate  a fake review classifier model  based on the framework previously described. The classifier has been trained and tested over the consumer electronics dataset previously scrapped and evaluated using ten fold cross-validation.
Also, we tackle the impact the user and review centric features may have on the performance, and the possible differences across different cities.
After analyzing the results obtained, a statistical evaluation has been carried out.

\subsection{Review centric features}

The technique which best performed for analyzing review centric textual features was \ac{TF-IDF} with bigrams, but it reached an F-Score below 60\%.
Previous research regarding the restaurant domain reached similar conclusions, stating that the text of the reviews were not a good indicator of reviews veracity~\cite{mukherjee1}. 
Nevertheless, the results obtained in the particular case of consumer electronics show that text information is not useful for fake classification in this domain. 
In this sense, bigram representations are considered as a fairly strong baseline~\cite{Wang2012,DBLP:journals/corr/JoulinGBM16}.
Obtaining a low performance with such baseline enforces the idea that the text does not serve as indicator for fake reviews.
Apart from using TF-IDF, we experimented with other techniques such as \ac{LDA} models or sentiment analysis, but they did not work well.
The results did not improve either with the use of Word2Vec models, as these performed similar to random guessing.

\subsection{User centric features}
\begin{figure*}[hbtp]
 \centering
 \includegraphics[width=0.4\textwidth]{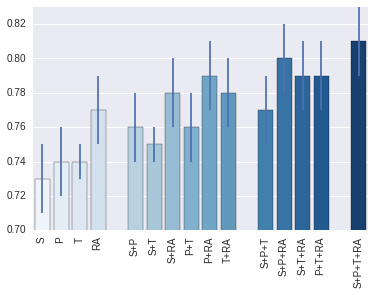}
 \caption{Performance of review centric features using an AdaBoost classifier}
 \label{fig:adaboostAll}
\end{figure*}

In the case of user centric features, results were clearly improved in comparison with review centric features. Regarding the experiments, we highlight the contributions of the different subdivisions of user centric features we defined in the proposed framework, which were \acf{S}, \acf{P}, \acf{T} and \acf{RA}. Table~\ref{table:results} shows the F-Scores obtained for each of the experiments done in every city, structuring the results based on city, type of features employed and classification algorithm (Logistic Regression, Decision Tree, Random Forest, Gaussian Naive Bayes, AdaBoost).
Additionally, results for the combination of all cities are shown.
Figure \ref{fig:adaboostAll} shows the F-Scores for several combinations of the proposed features aggregated by city, as obtained by AdaBoost, which is the classifier with the best performance. This way, it is easy to analyze the results considering several aspects and the following conclusions were drawn.

 \begin{table}[hbtp]
 \begin{center}
	\begin{tabular}{ l  l  l  l  l  l  l }
	\hline
	\textbf{City} & \textbf{Features} & \textbf{\acs{LR}} & \textbf{\acs{DT}} & \textbf{\acs{RF}} & \textbf{\acs{GNB}} & \textbf{\acs{AB}} \\
	\hline\hline
	All cities & S + P + T + RA & 0.77 & 0.80 & 0.80 & 0.71 & {\bf 0.81} \\
	\hline\hline
	New York    & S + P + T + RA & 0.79 & 0.81 & {\bf 0.82} & 0.72 & {\bf 0.82}  \\
	\hline
	Los Angeles & S + P + T + RA & 0.73 & 0.73 & 0.78 & 0.69 & {\bf 0.79}  \\
	\hline
    Miami	 & S + P + T + RA & 0.78 & 0.81 & 0.81 & 0.71 & {\bf 0.82 } \\
    \hline
	San Francisco & S + P + T + RA & 0.78 & 0.81 & 0.81 & 0.69 & {\bf 0.82}  \\
	
	\hline
    \multicolumn{2}{c}{Friedman rank} & 3.87 & 2.87 & 2.12 & 5 & 1.12 \\
	
	\end{tabular}
	
	\begin{tablenotes}
      \small
      \item Features legend: S (Social), P (Profile), T (Trust), RA (Reviewing Activity).
      \item Classifier legend: LR (Logistic Regression), DT (Decision Tree), RF (Random Forest), GNB (Gaussian Naive Bayes), AB (AdaBoost).
    \end{tablenotes}
    \caption{F-Score results}
	\label{table:results}
\end{center}
\end{table}

When comparing the impact of the different defined features, we can observe that \ac{RA} features are the most relevant and \ac{S} are the least ones. Nonetheless, our experiments show that social features improve the accuracy of the rest of feature types. Moreover, each combination of feature types consistently increases the performance. Thus, our initial hypothesis considering that social features could be effective is supported.

Finally, analyzing the results across the four different cities, it can be extracted that F-Scores are quite similar between New York, Miami and San Francisco, but Los Angeles achieves worse results if compared to the same pair of classifier-feature subset of any other city.
When taking into account all cities at the same time, it can be seen that the better results for individual cities are not surpassed.

\subsection{Combining review and user centric features}
From the previous results, it can be seen that there is a clear difference between using review or user centric features. However, the last step was an attempt of combining both kinds of features trying to improve the model. These experiments were not satisfactory in the case of textual models which incorporate large numbers of features such as \ac{TF-IDF} or Word2Vec, as those features were almost seen as noise without having any impact on the decisions taken by classifiers. Due to this, we also tried to incorporate a reduced subset of features extracted from those models but the results were the same.

In the case of adding sentiment analysis, POS features or other textual metrics to the user centric subset of features the results did neither improve as those features were also irrelevant for the classifiers. 
The  last  experiment  consisted  in  combining  the outputs of separate models built with user or review centric features following ensemble techniques.
Nevertheless, it was not effective in our dataset.
Some of the potential reasons of this are that fake reviewers have become more sophisticated, as well as that Yelp filter is exploiting mainly user centric features instead of linguistic ones based on their effectiveness as pointed out by~\cite{mukherjee1}.
Nevertheless, other research works~\cite{mukherjee1} have reported a slightly improvement with the combination of bigrams and behavioural features in the restaurants and hotels domain.
This discrepancy can be due to the fact the high competition in these domains and the economic incentives of fake reviews~\cite{luca2016fake,mayzlin2014promotional}.

The results of these experiments, that aim to exploit the information contained in the textual reviews, are made public online for the interested reader\footnote{\url{http://gsi.upm.es/~oaraque/FakeReviews/additionalmaterial.pdf}}.

\subsection{Statistical analysis}\label{sec:validation}
In order to compare the different learning models used in this work, a statistical test has been applied on the experimental data.
In particular, we have chosen the Friedman test \cite{demvsar2006statistical}, as this test is oriented to the comparison of several classifier methods on multiple datasets.

The Friedman test is based in a rank of each classification method in each dataset, where the best performing algorithm is assigned the rank of 1, the second best is assigned rank 2, etc.
Ties in this rank are resolved by the average of their ranks.
$r_{j}^{i}$ is the rank of the $j$-th classification methods on the $i$-th dataset, where $j \in \{1, 2, ..., k\}$ and $i \in \{1,2, ..., N\}$.
With the Friedman test, the average of the ranks $R_{j} = \frac{1}{N} \sum_{i} r_{i}^{j}$ is compared.
The null-hypothesis is defined as the situation where all the classification algorithms are equal, and so are their ranks $R_{j}$.
Then, the Friedman statistic with $k-1$ degrees of freedom is written as follows:
$$ \chi_{F}^{2} = \frac{12N}{k(k+1)} \left(  \sum_{j} R_{j}^{2} - \frac{k(k+1)^{2}}{4} \right)$$
Nonetheless, it is shown that there is a more useful statistic that is distributed according to the F-distribution, and has  $k-1$ and $(k-1)(N-1)$ degrees of freedom \cite{demvsar2006statistical}.
This statistic is referred to as the Friedman F, and is expressed as:
$$ F_{F} = \frac{(N-1)\chi_{F}^{2}}{N(k-1) - \chi_{F}^{2}} $$
If the null-hypothesis of the Friedman test is rejected, post-hoc tests can be conducted, complementing the statistical analysis.
In this work, we have conducted the Nemenyi and Holm tests, with the aim of gaining insight of the differences between the analyzed classifiers.

In the Nemenyi test, all classifiers are compared to each other.
In this way, the performance of a two classifiers is significantly different if their ranks differ, at least, the \textit{critical difference}.
The critical difference is computed with the following expression:
$$ CD = q_{\alpha} \sqrt{\frac{k(k+1)}{6N}} $$
For the Holm test, the \textit{z} value is computed, which allows to obtain a probability in from the normal distribution.
The z value that compares the $i$-th and $j$-th classifiers is computed as:
$$ z = \left. (R_i - R_j) \middle/ \sqrt{\frac{k(k+1)}{6N}} \right. $$
The Holm test compares the corresponding \textit{p}-values of each z value with a value $\alpha / (k -1)$ value in a step-down manner.
If the \textit{p}-value has a lower value than the modified $\alpha$ value, the null-hypothesis is rejected.
As soon as a certain null-hypothesis is rejected, all the remaining hypothesis are retained, that is, they can not be rejected.

The computation of the tests has been made as follows.
In relation to the \textbf{Friedman} test, the ranks have been computed.
The average ranks ($R_i$) are shown in Table~\ref{table:results}.
For all the calculations, the $\alpha$ value is set to 0.05.
Attending to those $R_i$ values, $\chi_{F}^{2} = 14.5$,  $F_{F} = 29.0$, and the critical value $F(k-1,(k-1)(N-1))= 3.26$.
Given that $F_{F} > F(4,12)$, the null-hypothesis is rejected, that is, not all the classifiers have similar performance, and post-hoc tests can be conducted.

The \textbf{Nemenyi} test is then computed, selecting $q_{\alpha} = 2.728 $, as indicated in \cite{demvsar2006statistical}.
With these values, the corresponding critical difference is $CD = 3.05$.
Given the $R_i$ values obtained in the Friedman test, it can be observed that the Nemenyi test points that the \acl{GNB} and the \acl{AB} classifiers are significally different.
Finally, the \textbf{Holm} test rejects the null-hypothesis for the \acl{AB} and \acl{RF} classifiers, with the following values: $p\vert_{\textrm{AB}} = 0.0005$, $\frac{\alpha}{i}\vert_{\textrm{AB}} = 0.0125$, $p\vert_{\textrm{RF}} = 0.01$ and $\frac{\alpha}{i}\vert_{\textrm{RF}} = 0.017$.

In conclusion, the statistical tests point that, between all the classifiers analyzed in this work, the \acl{AB} and \acl{RF} have the best performances in the datasets.
Attending to the Friedman ranks, we highlight the \acs{AB} performance in the studied problem.

\section{Conclusions}\label{sec:conclusions}
In this paper we have addressed fake review detection in the consumer electronics domain. We have proposed a feature framework oriented to analysis of social sites, and we have also developed a dataset which is made available\footnote{To obtain the dataset for research purposes, please write to \textit{o.araque@upm.es}.} for future research. Our framework is composed of two main types of features: review centric and user centric. Review centric features are only related to the text of the review. On the other hand, user centric features show how the user behaves within the site, and are subdivided into four groups: personal, social, reviewing activity and trust. 

As shown in the article, detecting fake reviews by just reading the reviews is a challenging task for both humans and computers, since  textual reviews do not usually provide fake signals to be detected. In contrast, features related to the user have been shown to be more effective. The most relevant ones are those coming from the reviewing activity. This is not surprising, since they can provide signals of abnormal reviews. The rest of features have also proven to have discrimination capability and every combination of feature types have improved the overall accuracy.

One of the conclusions of this paper is that fake reviews on the consumer electronics domain can be detected with a reasonably high F-Score using the features in the proposed \ac{F3} framework, reaching a maximum result of 82\% when using Random Forest or Ada Boost classifiers. Regarding the two main kinds of features described in the \ac{F3} framework, user centric features provide clearly better results than review centric features. This last method lead to F-Scores under 60\%, despite applying techniques such as \ac{TF-IDF} or neural learning models such as Word2vec. This fact had also been studied in other fields such as restaurant or hotel reviews~\cite{mukherjee1}, and it is reinforced by the difficulty of determining whether a review is fake or not by a human who reads it. Several experiments with the aim of combining both kinds of features were carried out including ensemble methods and dimensionality reduction techniques, but the results did not improve the user centric features benchmark. 

The use of social features, a subset of user centric features inside F3, had not been studied before in the fake review detection problem of any field. Interestingly, this set of features has brought great results reaching a maximum of 82\% F-Score in New York. As it has been said before, the most relevant features were the ones related to reviewing activity, but better results were achieved when combining different subsets of user centric features as each combination between the four subsets resulted into an increase of the F-Score. This fact indicates that the idea of dividing user centric reviews into four subsets has been successful, as features between subsets are independent and a classification system can take advantage of them when combined. 

Our dataset was compound of reviews from four of the most important USA cities, and results were quite similar between all of them. However, F-Scores in Los Angeles were a bit lower than in New York, Miami or San Francisco. With the final results obtained from each classifier incorporating all the features, we made a statistical analysis following the Friedman test, which shows that Ada Boost performs statistically better than the rest of used classifiers. 
 
Insights from this study can push forward the field of fake detection in social platforms.
The proposed framework is a first attempt to classify and organize the features used in this emerging research area.
As reported in this article, we have experimentally shown that using only the text of a review is not an effective approach, as other researchers have also shown before.
In light of this, common sense suggests that fake reviewers perform a fairly good job at disguising such invalid reviews, and even machine learning methods can be tricked.
Nevertheless, fake reviewers cannot hide their social network footprints and this can be a path for detecting them.
The proliferation of fake reviews and more recently fake news is a social problem that is overwhelming our society, so it is needed further research.
As future work, we propose to extend the proposed Fake Feature Framework (F3) on a general domain out of the e-commerce field.
Additionally, this framework can also be used in other tasks, such as toxic user activity detection or as  previously mentioned for fake news detection, which could involve complex semantic analysis.

\section*{Acknowledgements}
This work is supported by the Spanish Ministry of Economy and Competitiveness under the R\&D projects SEMOLA (TEC2015- 68284-R) and EmoSpaces (RTC-2016-5053-7), by the Regional Government of Madrid through the project MOSI-AGIL-CM (grant P2013/ICE-3019), the European Union under Trivalent (2020 RIA Action Grant No. 740934 under the call SEC-06-FCT-2016), and Spanish MINETAD (TSI-102600-2016-1).

\addcontentsline{toc}{chapter}{Bibliograpy}
\bibliographystyle{plain}
{
\small

}

\end{document}